# Condition directed Multi-domain Adversarial Learning for Loop Closure Detection *

1st Peng Yin
*State Key Laboratory of Robotics*
*Chinese Academy of Sciences*
Shenyang, China
yinpeng@sia.cn

2nd Yuqing He
*State Key Laboratory of Robotics*
*Chinese Academy of Sciences*
Shenyang, China
heyuqing@sia.cn

3rd Na Liu
*School of Mechatronics*
*Engineering and Automation*
*Shanghai University*
Shanghai, China
liuna_sia@shu.edu.cn

4th Jianda Han
*State Key Laboratory of Robotics*
*Chinese Academy of Sciences*
Shenyang, China
jdhan@sia.cn

5th Weiliang Xu
*Department of Mechanical Engineering*
*University of Auckland*
Auckland, New Zealand
p.xu@auckland.ac.nz

*Abstract*—Loop closure detection (LCD) is the key module in appearance based simultaneously localization and mapping (SLAM). However, in the real life, the appearance of visual inputs are usually affected by the illumination changes and texture changes under different weather conditions. Traditional methods in LCD usually rely on handcraft features, however, such methods are unable to capture the common descriptions under different weather conditions, such as rainy, foggy and sunny. Furthermore, traditional handcraft features could not capture the highly level understanding for the local scenes. In this paper, we proposed a novel condition directed multi-domain adversarial learning method, where we use the weather condition as the direction for feature inference. Based on the generative adversarial networks (GANs) and a classification networks, the proposed method could extract the high-level weather-invariant features directly from the raw data. The only labels required here are the weather condition of each visual input. Experiments are conducted in the GTAV game simulator, which could generated lifelike outdoor scenes under different weather conditions. The performance of LCD results shows that our method outperforms the state-of-arts significantly.

*Keywords—Loop Closure Detection; Generative Adversarial Networks; Multi-domain adversarial learning*

## I. INTRODUCTION

Loop closure detection (LCD) [1] or place recognition is the fundamental module in the simultaneous localization and mapping (SLAM) [2]. However, the LCD accuracy of current appearance based SLAM methods are usually affected by the illumination and texture changes under different weather conditions, such as sunny, foggy and rainy. Features under different weather conditions, show completely different characteristics, and in practice, we could not guarantee the visual inputs are obtained under same weather conditions. To improve the robustness to condition difference, Milfold et.al firstly proposes SeqSLAM [3], which use sequence matching instead of single scene recognition. The features used in the SeqSLAM are the hand-crafted and designed by experts with domain-specific knowledge. However, the handcraft features, such as the local feature descriptor SIFT [4], SURF [5], ORB [6] or global feature descriptor GIST [7], BRIEF [8] may fail by the affection of weather conditions. Furthermore, such features could not capture the local and global descriptions at the same time.

With the successful promotion of deep learning in the visual domain, such as object detection [9], scene segmentation [10] etc., many researchers try to apply the Deep neural networks (DNN) features into the LCD task. In Sunderhauf's work [11], they extract different layers to capture distinct property in LCD task. Usually, middle layers could extract the holistic geometry features, while deeper layers could capture the viewpoint-invariant features. Multi-layer DNN features could be used as a joint description for place recognition. However, since the networks are usually trained on the datasets under similar conditions [12], the accuracy of LCD could be affected in the real scenes with diverse conditions. Adversarial

Lowry et.al [13] assume that the differences caused by geometry and texture feature are relative small than the differences caused by season-to-season based appearance changes. To remove the season-to-season differences, Lowry proposed the season-invariant Change Removal method. The main idea of this approach is to remove the season based differences with a PCA based method. Then they use the re-generated season-invariant scenes for DNN feature extraction. Although the season features could be expelled, this approach view different conditions as the feature noise.

On the other hand, the networks module used in the above method are trained on specific visual task. Usually the data labels are carefully selected and evaluate with artificially calibrated ground truth. Thus with a given datasets in the real-life, constructing such labels is intractable. To improve the adaptability of the DNN features on the real-life place scene, Chen [14]proposed a training approach for CNN features. In this method, they manually divide the obtained scenes into several categories, and re-train the networks on the classification task. Then the semantic features are extracted from the last several layers same as in Sunderhauf's work. But it is uncertain to decide how many categories needed and how to estimate the category labels for each frame.

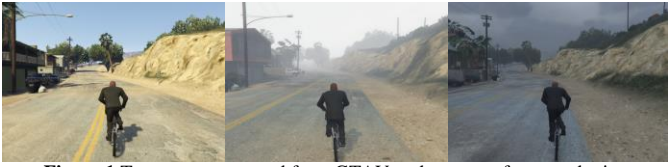

**Figure 1** Test scene generated from GTAV under sunny, foggy and rainy

The idea in this paper comes from the perception of human beings: as shown in the Figure 1, when people look at the scenes under different weather conditions, we won't ignore the weather effects on the scenes, instead we use such conditions as the direction for a better scene understanding. In this paper, we proposed a condition directed multi-domain adversarial learning (CMAL) for the appearance based loop closure detection. We use an unsupervised feature inference to extract efficient features for visual inputs, and use a classification module as the conditional directions for the better scene understanding. The main contribution of our method could be summarized as below:

- We propose a novel semi-supervised networks, which use the weather conditions as the direction for the feature inferences. This method could learning the weather invariant features directly from the raw images, with only the weather conditions for each frame. To our knowledge, we are the first to use the condition directed DNN features for the LCD task.

The rest of this paper is organized as below: firstly, Section II gives the recent works that has influenced our thought; secondly, Section III demonstrates the major modules in our proposed method; thirdly, to investigate the performance of our method, in the GTAV game simulator. We test the LCD accuracy under variant weather conditions, the experiment result shows that our method has outperformed other appearance based methods.

## II. PRIMARY

For better understanding for our proposed Stable-AFL, we first use introduce the generative adversarial networks (GANs) [15] based adversarial feature learning method. Then we briefly introduce the cycle-consistent adversarial networks [16], which method our proposed Stable-AFL is based on.

### A. GANs based Adversarial Learning

GANs is proposed by Goodfellow [15] et.al, as shown in Figure 1(a), this method uses a generator (or decoder) module $G$ to generate synthesis data from random latent code as real as possible, and a discriminator module $D$ to distinguish the real ones from synthesis ones as accurate as possible. The author uses a min-max value function to update the generator and discriminator modules,

$$\min_G \max_D V(D,G) = \\ E_{x \sim P_{data}(x)}\left[\log D(X)\right] + E_{z \sim P_z(z)}\left[\log\left(1-D(G(z))\right)\right], \quad (1)$$

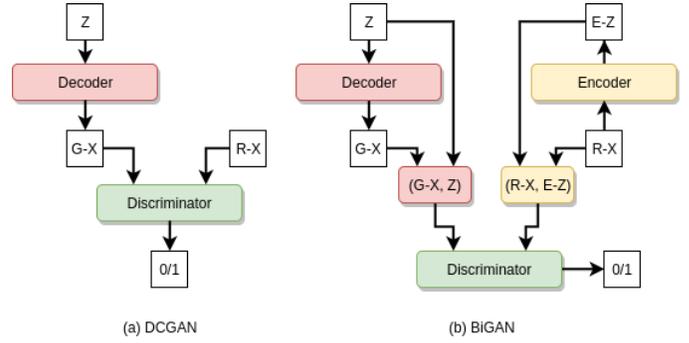

**Figure 2** The framework of GAN [17]

With the fixed generator, the value function could reach its optimal with the optimal discriminator $D^*$. $D^*$ satisfies [15] $D^*(x)=P_{data}(x)/(P_{data}(x)+P_g(x))$, where $P_g$ is the probability of data x in the synthesis data distribution. As proved by Goodfellow et.al, the optimal value function in Equation 1 could be written by,

$$C(G) = \max_D V(G,D) = 2 \cdot D_{JS}\left(p_{data} \| p_g\right) - 2\log 2 \quad (2),$$

where $D_{JS}$ is the Jensen-Shannon divergence (JSD). Since the JSD is non-negative and zero only when the real data distribution is equal to the synthesis data distribution. However, GAN based method could not encode the latent codes from the raw data. Based on the framework of GANs, Donahue [18] proposed a bidirectional generative adversarial networks (BiGANs) as shown in Figure 1(b). This method uses an additional encoder to mapping data into the latent code domain, and the discriminator is updated to distinguish the joint distribution of data and relative latent code. In our previous work, we have used BiGANs to extract the adversarial feature for the visual inputs. However, in the multi-condition case, BiGANs may also encode the conditions into the latent codes, which may cause diverse difference in the latent codes under different conditions.

### B. Maintaining the Integrity of the Specifications

Recently in the GANs family, there are some successful methods in image style transforming [19] [20] [16]. Kim et.al [19] use the GANs to discover cross-domain relations, as shown in Figure 3. This method uses two branches of encoder-decoder modules for two domain transforming. For domain $A$, they first use an encoder-decoder module $G_{AB}$ to map image $A$ into synthesis image $G$-$B$, then the synthesis data is reconstruct into domain $A$ with another encoder-decoder module $G_{BA}$. And the operations are same in another branch. The domain transferring is achieved by using discriminator A and B for distinguish synthesis data and the real ones, using encoder-decoder module to generate the synthesis data as real as possible. Ideally, the equality of $G_{BA}*G_{AB}(X_A)=X_A$ should be always satisfied. So there usually exists reconstruction losses $d(G_{BA}*G_{AB}(X_A), X_A)$ and $d(G_{AB}*G_{BA}(X_B), X_B)$ for both domain $A$ and $B$.

However such method could only be applied in two domain transferring task. In this paper, the proposed method could use the conditions as the direction for the multi-domain adversarial feature learning.

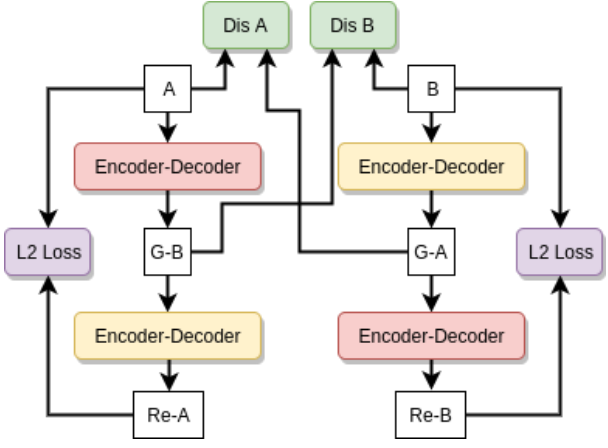

**Figure 3** Cross Domain Transferring.

## III. PROPOSED METHOD

In this section, we will firstly explain why the conditions could be used to direct the adversarial feature learning, and then explain the detail of CMAL method. Secondly, we will explain how the CMAL features could be used for the LCD task.

### A. Condition directed Adversarial Feature Learning

As shown in the Figure 4, in the multi-weather based scenes, each image is the combination of geometry features and the relative weather conditions. The desired LCD features should only capture the geometry differences and ignore the weather conditions. The main challenge for LCD intended feature extraction is to efficient separate the geometry features from the weather conditions. Instead of regarding the conditions as the feature noise, we view outer condition as the directions for multi-domain adversarial learning.

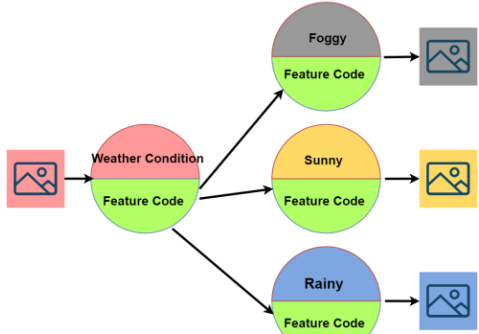

**Figure 4** Multi-domain Transferring

With the given data $X$, we assume that the real data distribution is combined by the independent condition probability distribution $P_{Con}$ and the unknown geometry probability distribution $P_{Geo}$,

$$P_A(Z, C_A | X) = P_{Geo}(Z|X) \cdot P_{Con}(C_A|X), \\ P_B(Z, C_A | X) = P_{Geo}(Z|X) \cdot P_{Con}(C_B|X). \quad (3)$$

where $Z$ is the geometry code and $C_x$ is the estimated conditions. The conditional probability could be obtained by a convolution neural networks (CNNs) based classification module.

Assume we have a encoder module $P_E(Z|C_x,X)$ that could extract the condition depended geometry feature. Then with the weather conditions $P_{Con}$, the reconstructed data distribution in domain A ($P_{RA}$) and domain B ($P_{RB}$) could be obtained by,

$$P_{RA}(Z, C_A|X) = P_E(Z|C_A, X) \cdot P_{Con}(C_A|X), \\ P_{RB}(Z, C_B|X) = P_E(Z|C_B, X) \cdot P_{Con}(C_B|X). \quad (4)$$

On the other side, we could also generate the synthesis data in domain A ($P_{GA}$) and domain B ($P_{GB}$) with a random noise distribution $P_Z(Z)$ as in the original GANs,

$$P_{GA}(Z, C_A|X) = P_Z(Z) \cdot P_{Con}(C_A|X), \\ P_{GB}(Z, C_B|X) = P_Z(Z) \cdot P_{Con}(C_B|X). \quad (5)$$

For all the reconstructed data distribution and generate data distribution, the relative distance to the real data distribution could be pull closer by using the min-max value updating approach as in the original GANs. Ideally, after well trained on the given datasets, the data distributions should follow the equation below,

$$P_A(Z, C_A|X) = P_{RA}(Z, C_A|X) = P_{GA}(Z, C_A|X), \\ P_B(Z, C_B|X) = P_{RB}(Z, C_B|X) = P_{GB}(Z, C_B|X), \quad (6)$$

and by combine Equation 3~5 in Equation 6, we could obtain,

$$P_{Geo}(Z|X) = P_E(Z|C_A, X) = P_Z(Z) \\ P_{Geo}(Z|X) = P_E(Z|C_B, X) = P_Z(Z) \quad (7)$$

Because the random noise $P_Z$ follows the same distribution, Equation 7 could be rewritten by,

$$P_{Geo}(Z|X) = P_E(Z|C_A, X) = P_E(Z|C_B, X) = P_E(Z|X) \quad (8)$$

which means the latent code generated from the encoder module $En$ could ideally separate the geometry features from the weather conditions.

The above proof is the core for our proposed CMAL method. With this property, we could use the weather conditions as the directions for multi-domain feature inference.

### B. CMAL

Based on the analysis in the previous section, to enable the condition directed multi-domain adversarial learning, we construct our networks with four core modules:

- A classification module $C$, classifies the raw data into different weather conditions;
- An encoder module $En$, extracts the geometry features from the raw data based on relative condition;
- A decoder module $De$, generate synthesis data based on geometry features and the relative condition;
- A discriminator module $D$, distinguishes the real ones from the synthesis ones based on relative weather condition.

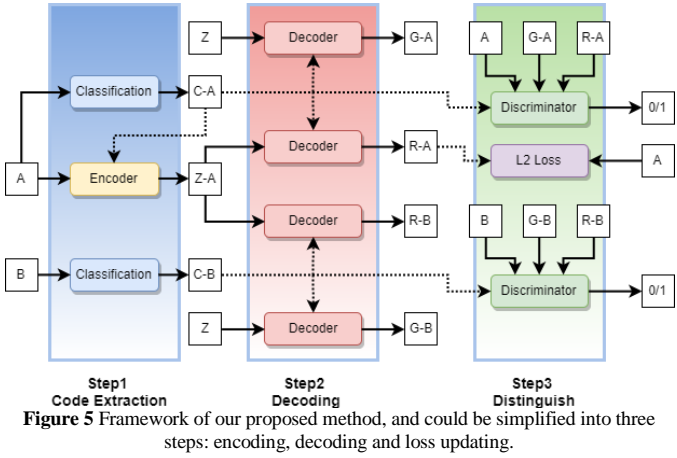

**Figure 5** Framework of our proposed method, and could be simplified into three steps: encoding, decoding and loss updating.

The framework is given in Figure 5, the procedure could be divided into three step.

1. **Encoding**: With the given two data A and B from two random domains, we first classify their categories as *C-A* and *C-B* with the supervised classification module. Then the geometry features *Z-A* is extracted by the encoder module under the condition *C-A*.
2. **Decoding**: With the geometry feature *Z-A*, we reconstruct the image into A *(R-A)* and B (R-B) domain under different weather conditions. Besides, we also generate the synthesis data in the domain A *(G-A)* and domain B *(G-B)* with the geometry feature Z from the random noise.
3. **Loss Updating**: In the final step, we use the conditional discriminator to separate the real ones from the synthesis ones. On the other hand, we also use a L2 loss function to enhance the reconstruction ability of encoder-decoder module.

To pull the reconstructed and generated data distribution closer to the real data distribution, here we use the discriminator module *D* and the encoder-decoder based generator to play the min-max value updating function same as in the original GANs,

The only difference in our value functions is that we use the conditional discriminator here. The value function for branch A is given by,

$$\min_D \max_{De,En} V_A(D, De, En) = E_{real} + E_{generate} + E_{reconstructed}$$
$$= E_{x \sim P_A(x)} \left[ \log D(x|C_A) \right]$$
$$+ E_{z \sim P_Z(z)} \left[ \log \left(1 - D(De(z|C_A)|C_A)\right) \right] \quad (9)$$
$$+ E_{x \sim P_A(x)} \left[ \log \left(1 - D(De(En(x|C_A)|C_A)|C_A)\right) \right],$$

where the encoder, decoder and discriminator module are all conducted under condition *A*. This min-max network updating will enforce the data distribution follow the first equation in Equation 6. And for the second equation in Equation 6, we have,

$$\min_D \max_{De,En} V_B(D, De, En) = E_{real} + E_{generate} + E_{reconstructed}$$
$$= E_{x \sim P_B(x)} \left[ \log D(x|C_B) \right]$$
$$+ E_{z \sim P_Z(z)} \left[ \log \left(1 - D(De(z|C_B)|C_B)\right) \right] \quad (10)$$
$$+ E_{x \sim P_B(x)} \left[ \log \left(1 - D(De(En(x|C_B)|C_B)|C_B)\right) \right],$$

On the other hand, the higher reconstruction ability means the latent code generated from the encoder could carry more geometry details. To enhance this ability, we use a *L2* loss term for the real data *A* and reconstructed data *R-A*,

$$\min_{En,De} L_R(En, De) = Loss\left(x_A, De(En(x|C_A)|C_A)\right). \quad (11)$$

With all the above loss functions, the joint value function is given by,

$$V(D, De, En) =$$
$$V_A(D, De, En) + V_B(D, De, En) + L_R(En, De). \quad (12)$$

### C. Sequence matching based LCD

Same as in the SeqSLAM [3], our method also relies on the sequence matching approach for the Loop closure detection. In the original SeqSLAM, there exists three steps: firstly given the pre-stored sequences $L_p$ and test sequences $L_t$, the difference matrix is calculated based on the sum of absolute difference (SAD) between $L_p$ and $L_t$; secondly, a local enhanced operation is taken to avoid the sequence searching stocked into local high similar area; finally, to find the best matches, different routes are visited and the only the routes with lower sum of differences are selected as the matches.

Here we use the difference in latent codes to replace the original SAD. With the extracted latent codes, the distance of two frames could be estimated by the Euclidean distance,

$$Diff(v_i, v_j) = \|v_i - v_j\|^2 \quad (13)$$

where $v_i$ is the encoded latent-code from the real data.

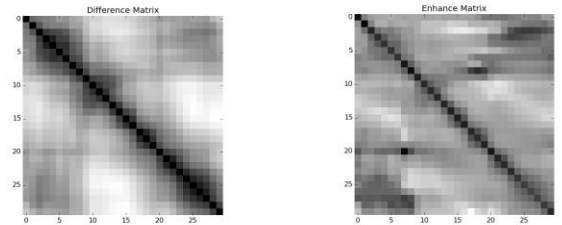

**Figure 6** Difference Matrix and Enhanced Matrix. The left figure is the difference matrix based on the Equation 13. The right figure is the enhanced difference matrix based on Equation 14.

The extracted difference matrix is shown in the left side of Figure 6, however, the differences are blurring, and the routes searching will be uncertain. To avoid this problem, a local data enhancement operation is applied,

$$\hat{D}_i = \frac{D_i - \bar{D}_l}{\sigma_l} \quad (14)$$

where $\bar{D}_l$ is the local mean and $\sigma_l$ is the local standard deviation in the nearest difference values of $D_i$. The right side of Figure 6 shows the result of local enhancement.

In the final step, to recognize the best match, a space window $M$ with the recent image difference vectors is used for searching the best sequence matches:

$$M = \left[ \hat{D}^{T-d_s}, \hat{D}^{T-d_s+1} \ldots \hat{D}^T \right] \quad (15)$$

where $D^T$ is the column vector as shown in the right blue box in Figure 7, which represent the difference vector of test frame at timestamp $T$ with the pre-stored sequences. $d_s$ is the time length for watch back searching. The difference score $S$ is then calculated for routes with different velocity,

$$S = \sum_{t=T-d_s}^{T} D_k^t, k = s - V(d_s - T + t) \quad (16)$$

where $s$ is the relative position in train frame, $V$ is the potential velocity proportion of test sequence and train sequence. The loop closures are estimated with the difference score $S$ small than a given threshold.

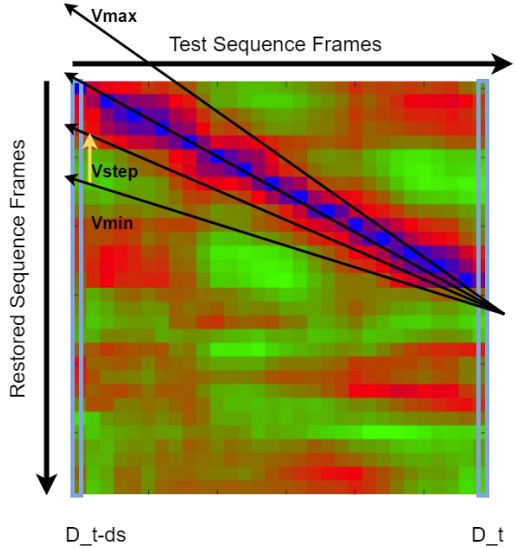

**Figure 7** Sequence frame matching.

## IV. EXPERIMEN

To investigate the performance of our proposed method in the multi condition based LCD task, we test the LCD accuracy under sunny, rainy and foggy weather conditions in the GTAV game simulator. The experiment is tested with a single NVidia Titan X card with 64G RAM on the Ubuntu 14.04 system. For the training datasets, we collect 10,000 images for each weather condition in the north mountain area in the GTAV.

The testing datasets are completely different from the training data, we pick up two fixed routes and generate the test sequence under different weather conditions. And we check the LCD accuracy under *sunny-foggy*, *foggy-rainy* and *rainy-sunny* matches. In this experiment, we use different features under the same sequence matching approach. The features we used here are SAD (as in the original SeqSLAM), DNN based features (where the DNN features are extracted in the fully connected layers of VGG 16 networks with a pre-trained model on the ImageNet datasets), BiGAN based features and our proposed CMAL features. The parameters used in sequence matching are listed in TABLE I, where $d_s$, $V_{min}$, $V_{max}$, and $V_{step}$ are the parameters shown in Figure 7. And $D_{thresh}$ is the distance threshold to judge whether the matching are satisfied.

TABLE I
PARAMETERS USED IN ENHANCED SEQSLAM

| Parameter | Description |
|---|---|
| $d_s$ | The length of watch back trajectory, in this experiment, we set $d_s$=10 |
| $V_{min}$ | Minimum trajectory velocity proportion, here $V_{min}$=0.8 |
| $V_{max}$ | Minimum trajectory velocity proportion, here $V_{min}$=1.1 |
| $V_{step}$ | The step forward value for velocity proportion, $V_{step}$=0.1 |
| $D_{thresh}$ | The frame distance of matched frames to decide whether matchings are satisfied, $D_{thresh}$=20 |

### A. Measurement Metrics

To measure the LCD accuracy of different methods, we make qualitative analysis with PRC (Precision-Recall curve) and AUC (area under the Receiver operating characteristic (ROC)); for the quantitative analysis, we use the recall at 100% perception in the PRC to measure LCD accuracy. Here, for the matched pairs, if the distance between ground truth position and estimated one is within $D_{thresh}$, then the pairs are regarded as true positive (TP), else will be regarded as false positive (FP); on the other side, the pairs erroneously discarded by the match score are regarded as false negative (FN), and the ones of actually no-matched pairs are regarded as the true negative (TN). Thus the precision and recall are then obtained by,

$$\Pr ecision = \frac{TP}{TP+FP}$$
$$\operatorname{Re} call = \frac{TP}{TP+FN} \quad (17)$$

The AUC score is the size of covered ROC area, and the ROC curve is created by plotting the true positive rate (TPR) against the false positive rate (FPR) at various threshold settings, which are obtained by,

$$TPR = \frac{TP}{TP+FN}$$
$$FPR = \frac{FP}{TN+FP} \quad (18)$$

### B. Accuracy Analysis

Figure 8~9 show the PR curves of different methods on route1 and route2. As we can see, under variant weather conditions, the PR curve of our proposed CMAL features is better than the SAD, DNN and BiGAN based feature in the Sequence matching based LCD task.

Figure 10 gives a more directly view with the AUC indexes on the two routes, where the higher the index the more robust in the LCD task. In general, CMAL is also better than the DNN, BiGAN features, but slight better than the SAD feature.

Table II gives the quantitative demonstration with the recall indexes under precision rate at 100%. In overall, the CMAL features is better than the DNN and BiGAN based method, but is worse than the SAD features.

TABLE II
RECALL AT 100% PRECISION

| Parameter | Route1 | | | Route2 | | |
|---|---|---|---|---|---|---|
| | Foggy Rain | Foggy Sunny | Rain Sunny | Foggy Rain | Foggy Sunny | Rain Sunny |
| SAD | 34.8% | 29.7% | 65.6% | 11.0% | 50.8% | 73.1% |
| CMAL | 14.1% | 48.3% | 28.7% | 31.9% | 10.1% | 22.4% |
| DNN | 0.1% | 4.5% | 0.4% | 0.8% | 2.1% | 0.6% |
| BiGAN | 31.4% | 16.4% | 6.4% | 10.6% | 2.6% | 5.4% |

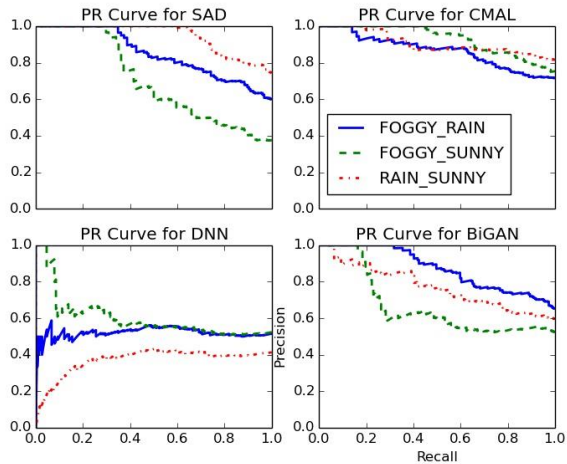

Figure 8 PR Curves for Route 1

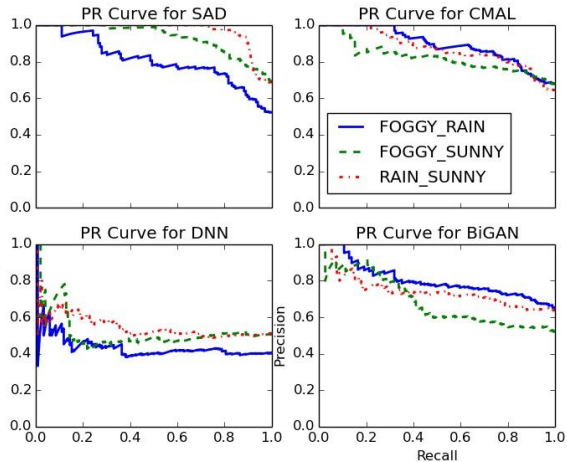

Figure 9 PR curve for the Route 2

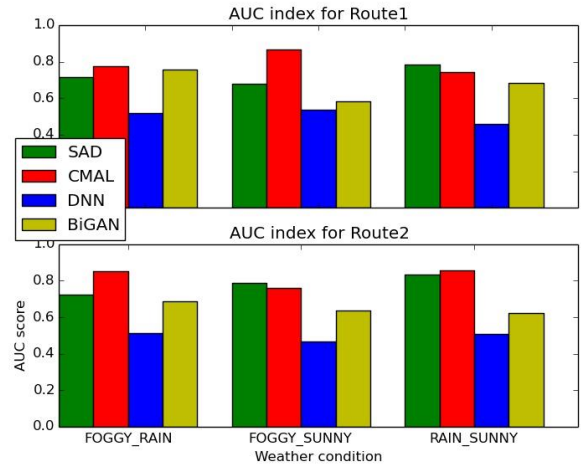

Figure 10 AUC indexes for LCD

Consider the above the results, we can see the proposed CMAL based feature is better than the DNN features extracted from VGG16, and the BiGAN based feature, i.e. the condition directed module is worked for better geometry feature inference. However, the SAD feature is still holding a relative high LCD accuracy. We think this is because the frames are generated the GTAV, the geometry is fixed in the game simulator, and since the SAD relies on the grayscale inputs, the weather conditions will have limited effects on the grayscale based geometry difference. The relative matching details could be found on our YouTube site[1].

### C. Image Reconstruction

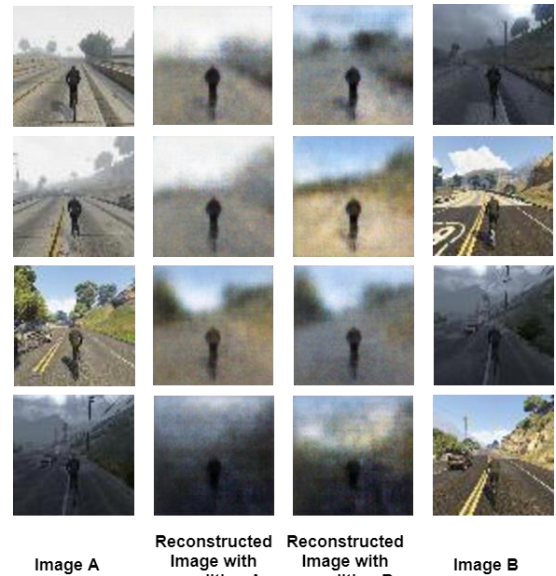

Figure 11 Samples of Map Reconstruction. The left column are the images in original domain A, the most right column are the target domain B. With the same latent code extracted form domain A, the second and third columns are the reconstructed images with condition A and condition B.

---

[1] https://youtu.be/7nusFt8T1a0

Figure 11 shows the reconstructed samples in our proposed method. The first column shows the geometry features are extracted from the random domain *A*. Then with such features, column two reconstruct the image into the *A* domain with weather condition in *A*, and column three reconstructs the image into the *B* domain with weather condition *B*. The last column shows the images in the random domain *B*. As we can see the reconstructed images in domain *A* and *B* all capture the similar geometry and with different weather conditions, just following what we designed for.

*D. Runtime and storage Analysis*

The mixture features are saved as a 1024 vector in the float32 format, so each code is occupied for 1024*4B=4KB, which is relative easier for the normal storage saving. For the runtime analysis of feature inference, the average feature inference time is shown in Table III. As we can see, the average feature inference time in the CMAL is around 3.5ms, which is very fast for the visual inputs recognition.

With the above property in storage and runtime, the proposed method could be easily plugged on any kinds of mobile robots for the real time long term navigation task.

TABLE III
FEATURE INFERENCE TIME PER FRAME (MILLISECOND)

| Parameter | Route1 | | | Route2 | | |
|---|---|---|---|---|---|---|
| | Foggy Rain | Foggy Sunny | Rain Sunny | Foggy Rain | Foggy Sunny | Rain Sunny |
| SAD | 3.2 | 3.2 | 3.2 | 3.3 | 3.4 | 3.3 |
| CMAL | 3.4 | 3.5 | 3.6 | 3.4 | 3.4 | 3.5 |

V. COMCLUSION

In this paper, we proposed a novel condition directed multi-domain adversarial learning (CMAL) method. Based on generative adversarial networks (GANs) and a classification networks, we design a semi-supervised feature extraction method. The key idea of our work is to regard the outer conditions as the directions, instead of the feature noise. In the experiment part, we generate sequence frames on different routes under sunny, foggy and rainy weather conditions in the GTAV game simulator. The results show that, the proposed CMAL features could do better in the geometry feature extraction for the LCD task.

We could easily extent the CMAL method to more complex conditions, such as time in a day, season. In the future work, we will investigate the CMAL encoding ability to use more complex conditions as the directions.